\documentclass{article}

\usepackage{microtype}
\usepackage{graphicx}
\usepackage{subfigure}
\usepackage{booktabs} 
\usepackage{ltablex} 
\usepackage{colortbl}
\usepackage{ragged2e} 
\usepackage{caption}  
\usepackage{multirow}

\usepackage{hyperref}

\usepackage[accepted]{icml2025}

\usepackage{amsmath}
\usepackage{amssymb}
\usepackage{mathtools}
\usepackage{amsthm}

\usepackage[capitalize,noabbrev]{cleveref}

\theoremstyle{plain}

\theoremstyle{definition}

\theoremstyle{remark}

\usepackage[textsize=tiny]{todonotes}

\icmltitlerunning{Configurable Preference Tuning with Rubric-Guided Synthetic Data}

\begin{document}

\twocolumn[
\icmltitle{Configurable Preference Tuning with Rubric-Guided Synthetic Data}



\icmlsetsymbol{equal}{*}

\begin{icmlauthorlist}
\icmlauthor{Víctor Gallego}{yyy}
\end{icmlauthorlist}

\icmlaffiliation{yyy}{Komorebi AI Technologies, Madrid, Spain}

\icmlcorrespondingauthor{Víctor Gallego}{victor.gallego@komorebi.ai}

\icmlkeywords{direct preference optimization, alignment, synthetic data, preference tuning}

\vskip 0.3in
]



\printAffiliationsAndNotice{}  

\begin{abstract}
Models of human feedback for AI alignment, such as those underpinning Direct Preference Optimization (DPO), often bake in a singular, static set of preferences, limiting adaptability. This paper challenges the assumption of monolithic preferences by introducing Configurable Preference Tuning (CPT), a novel framework for endowing language models with the ability to dynamically adjust their behavior based on explicit, human-interpretable directives. CPT leverages synthetically generated preference data, conditioned on system prompts derived from structured, fine-grained rubrics that define desired attributes like writing style. By fine-tuning with these rubric-guided preferences, the LLM learns to modulate its outputs at inference time in response to the system prompt, without retraining. This approach not only offers fine-grained control but also provides a mechanism for modeling more nuanced and context-dependent human feedback.

Several experimental artifacts, such as training code, generated datasets and fine-tuned models are released at \href{https://github.com/vicgalle/configurable-preference-tuning}{github.com/vicgalle/configurable-preference-tuning}
\end{abstract}

\section{Introduction}
The remarkable progress of Large Language Models (LLMs) has opened up a wide array of applications. However, aligning these models with desired human preferences, behaviors, and safety protocols remains a significant challenge. Techniques like Reinforcement Learning from Human Feedback (RLHF) \cite{ziegler2019fine,christiano2017deep,ouyang2022training} and Direct Preference Optimization (DPO) \cite{rafailov2024direct} have shown success in steering LLMs towards preferred responses. However, a critical, often implicit, assumption underpins many existing human feedback models: the notion of a singular, static, and monolithic set of preferences. Human preferences are rarely monolithic; they are dynamic, context-dependent, and multifaceted, influenced by factors ranging from individual user needs and cultural norms to evolving ethical considerations and task-specific requirements. Current models, by "baking in" an averaged or aggregated preference profile during fine-tuning, often lack the adaptability to reflect this richness. This inflexibility means that altering an LLM's behavior—for instance, to adjust its writing style, modify its safety strictures for different environments, or cater to diverse user cohorts—typically necessitates resource-intensive retraining or further fine-tuning. Such limitations hinder the development of truly robust, interpretable, and adaptable AI systems capable of genuinely understanding and responding to the spectrum of human intentions.

This paper directly addresses this limitation by challenging the assumption of monolithic preferences. We introduce Configurable Preference Tuning (CPT), a novel framework that endows LLMs with the ability to dynamically adjust their behavior at inference time based on explicit, human-interpretable directives. CPT leverages synthetically generated preference data conditioned on system prompts that are derived from structured, fine-grained rubrics. These rubrics may define desired attributes—such as stylistic nuances, safety levels, or persona adherence—along various dimensions. By fine-tuning an LLM with these rubric-guided preference pairs using a DPO-style objective, the model learns to modulate its outputs in response to the corresponding system prompt, without requiring retraining for each new configuration.

Our contribution offers a pathway towards more granular, transparent, and controllable alignment. It moves beyond a single "one-size-fits-all" preference model, allowing for the explicit specification and operationalization of diverse behavioral configurations.  We demonstrate that CPT enables fine-grained control contributing to the development of more robustly aligned AI systems that can better reflect the multifaceted nature of human feedback.

\subsection{Related Work}
The challenge of moving beyond a single, averaged preference model in LLMs has spurred growing interest in personalized Reinforcement Learning from Human Feedback (RLHF). Broadly, approaches to specialize LLM behavior can be seen through different lenses. Some methods aim to derive a single policy that represents a compromise or aggregation of diverse user preferences \cite{dumoulin2023density,conitzer2024social}. While these improve upon a simple average, they may not fully cater to specific, nuanced individual needs.

Closer to our work are approaches designed for downstream specialization of a policy or its underlying reward model to a particular user, persona, or specified context. Some methods learn a direct mapping from user-specific information (e.g., interaction history, user IDs, or textual descriptions) to tailored reward signals or policy adjustments. For instance, \cite{poddar2024personalizing} use variational preference learning to encode user rankings into a latent variable conditioning the reward model. \cite{li2024personalized} compute user embeddings to condition a base LLM via soft prompting in their P-RLHF framework. These methods often rely on inferring latent representations of user preferences, which, while powerful, may lack the direct interpretability and explicit controllability offered by rubric-based specifications. \cite{gallego2024configurable} enhances DPO for language models by allowing flexible safety configurations via system prompts without hard-coding behaviors, but doesn't account for non-binary preference levels.

Another line of research, exemplified by the work of \cite{barreto2025capturing} on Reward Feature Models (RFMs) and related approaches \cite{chen2024modeling, go2023compositional}, focuses on learning a set of underlying \emph{reward features} from context-response pairs. User-specific preferences are then modeled by learning a set of weights for these features, often through adaptation with a few examples from the target user. The work of \cite{barreto2025capturing} demonstrate that an RFM can be trained on pairwise comparisons, resulting in reward features that are linearly combined with user-specific weights $w_h$ to represent $p(y \succ y'|x, h)$, enabling fast adaptation to new users by learning these weights. Their approach effectively aims to discover latent criteria from data and then allows users to re-weight these criteria.

Our Configurable Preference Tuning (CPT) framework shares the overarching goal with these latter approaches: enabling fine-grained, user-directed control over LLM outputs. However, CPT diverges in its mechanism for specifying and learning these configurations. Rather than learning latent reward features from general preference data and then adapting weights for individual users ($h$), CPT utilizes \textit{explicitly defined rubrics} as the source of stylistic dimensions. These rubrics, paired with target scores, guide a teacher model to generate synthetic preference data. The student model is then fine-tuned using Direct Preference Optimization (DPO) to respond to \textit{system prompts} ($s$) which are concise summaries of these rubric-score combinations. Thus, while RFMs learn to adapt $p(y \succ y'|x, h)$ by inferring $w_h$ for learned features $\phi_\theta(x, y)$, CPT directly learns $p(y \succ y'|x, s)$ where $s$ is a declarative instruction about the desired style, operationalized through rubric-guided synthetic data. The ``features'' in CPT are implicitly defined by the rubric criteria and are selected/modulated by the system prompt, rather than being learned end-to-end as in RFM. This allows CPT to integrate rich, human-understandable stylistic desiderata directly into the fine-tuning process.

\section{Configurable Preference Tuning}

Our framework aims to learn a preference model $p(y_w \succ y_l | x, s)$, where $y_w$ is the preferred (winner) response and $y_l$ is the dispreferred (loser) response to a user prompt $x$, given a system prompt $s$ that expresses the desired configuration. This contrasts with standard preference modeling $p(y_w \succ y_l | x)$, which lacks the conditioning on $s$.

\subsection{Synthetic Preference Data Generation}

The core of CPT lies in its method for generating diverse, configurable preference data without requiring new human annotations for each desired configuration. This process involves the following steps:

\begin{enumerate}
\item \textbf{Rubric Definition ($\mathcal{R}$):} We define a set of rubrics, $\{\mathcal{R}_i\}$, each detailing specific attributes or styles for LLM responses. For instance, a rubric might specify criteria for ``formality,'' ``creativity,'' ``safety level,'' or ``adherence to a persona.'' Each rubric implicitly defines an axis of variation. Two examples of the rubrics we used in the experimental section can be found in Tables \ref{tab:chaos_cinema} and \ref{tab:rococo_reviewer} in the Appendix.

\item \textbf{Score-Conditioned Generation:} For each rubric $\mathcal{R}$ and user prompt $x$, we can prompt a capable teacher LLM to generate responses that achieve different target scores or levels (e.g., \texttt{low score}, \texttt{moderate score}, \texttt{high score}) with respect to that rubric. This is achieved using an augmented prompt $\phi(x, \mathcal{R}, \texttt{score})$, which instructs the teacher model, as seen in Table \ref{tab:teacher_prompt}. This allows us to sample responses $y \sim p(y|\phi(x, \mathcal{R}, \texttt{score}))$ aligned with different rubrics $\mathcal{R}$ and score levels.

\begin{table}[!h]
\caption{Prompt for generating responses aligned with $\mathcal{R}$ and \texttt{score}.}
\label{tab:teacher_prompt}
\vskip 0.15in
\begin{center}
\footnotesize
\begin{tabular}{c}
\toprule
\begin{minipage}[t]{0.8\columnwidth}%
\texttt{Your response will be evaluated using the following rubric $\lbrace \mathcal{R}\rbrace $. Given the following task: $\lbrace x\rbrace $, generate a response that achieves $\lbrace \text{score}\rbrace $ in the previous rubric.} 
\end{minipage}\tabularnewline
\bottomrule
\end{tabular}

\end{center}
\vskip -0.1in
\end{table}

\item \textbf{System Prompt Synthesis ($s$):} For each rubric $\mathcal{R}$ and target $\texttt{score}$, we generate a concise system prompt $s = \text{summarize}(\mathcal{R}, \texttt{score})$. This system prompt is a natural language instruction that encapsulates the essence of achieving $\texttt{score}$ under rubric $\mathcal{R}$, and is obtained by prompting the same teacher models to summarize the rubrics into a brief instruction of two to three sentences. Table \ref{tab:system_prompts_alt} shows several examples of summarized system prompts.

\item \textbf{Constructing Preference Pairs:} To create DPO training instances, we select a rubric $\mathcal{R}$ and two distinct target scores, $\texttt{score}_1$ and $\texttt{score}_2$. We then generate corresponding responses $y_1$ and $y_2$ using the teacher model. We also generate their associated system prompts $s_1$ and $s_2$ according to the previous step.

This yields two preference tuples for our training dataset:
\begin{itemize}
\item The first tuple conditions on $s_1$: Given user prompt $x$ and system prompt $s_1$ (which desires behavior aligned with $\text{score}_1$), $y_1$ is preferred over $y_2$. The DPO training sample is effectively (prompt: $(s_1, x)$, chosen: $y_1$, rejected: $y_2$).
\item The second tuple conditions on $s_2$: Given user prompt $x$ and system prompt $s_2$ (which desires behavior aligned with $\text{score}_2$), $y_2$ is preferred over $y_1$. The DPO training sample is (prompt: $(s_2, x)$, chosen: $y_2$, rejected: $y_1$).
\end{itemize}

This construction is crucial as it teaches the student LLM to switch its preference based on the provided system prompt $s$, using the same underlying pair of generated responses $(y_1, y_2)$. The end result of this process is a preference dataset $\mathcal{D} = \lbrace (s, x, y_w, y_l)_i \rbrace_{i=1}^N$.
\end{enumerate}

\subsection{Illustrative Example: Stylistic Control}

Let $x$ be \texttt{Generate a movie review for a movie you liked}. Let $\mathcal{R}$ be the rubric from Table \ref{tab:chaos_cinema} that emphasizes texts written in an unconventional style.

\begin{itemize}
\item $\texttt{score}_1 = \texttt{extremely high score}$ .

$s_1$ =  ``Generate a text that is fragmented, illogical, and filled with unexpected connections, embracing absurdity and subverting conventional expectations of language and form.''. Teacher model generates $y_1$.

\item $\texttt{score}_2 = \texttt{low score}$.

$s_2$: ``Write in a clear, concise, and completely conventional style, adhering strictly to established norms of grammar, syntax, and logical coherence.''. Teacher generates $y_2$ (a review written using standard language).
\end{itemize}

The CPT dataset would include:
\begin{enumerate}
\item For system prompt $s_1$: $(s_1, x, y_1, y_2)$ indicating $y_1 \succ y_2$.
\item For system prompt $s_2$: $(s_2, x, y_2, y_1)$ indicating $y_2 \succ y_1$.
\end{enumerate}

\subsection{Training with DPO}
Once we have the preference dataset, we can use it to align any LLM (the student) to these diverse sets of preferences. The student LLM is fine-tuned using DPO \cite{rafailov2024direct}. The DPO loss function aims to increase the likelihood of the preferred response and decrease the likelihood of the rejected response, conditioned on both the original user prompt 
$x$ and the generated system prompt $s$. The input to the model during DPO training is effectively a concatenation or structured combination of $s$ and $x$, writing the DPO loss function as $\mathcal{L}_{\text{DPO}}(\pi_\theta; \pi_{\text{ref}}) =  -\mathbb{E}_{(s,x,y_w,y_l) \sim \mathcal{D}} \left[ \log \sigma \left( \beta \log \frac{\pi_\theta(y_w | s, x)}{\pi_{\text{ref}}(y_w | s, x)} - \beta \log \frac{\pi_\theta(y_l | s, x)}{\pi_{\text{ref}}(y_l | s, x)} \right) \right]$. 
This process distills the nuanced, rubric-guided behaviors into the student model, making them controllable via 
$s$ at inference time.

\section{Experiments}

To validate the efficacy of Configurable Preference Tuning (CPT), we conducted a series of experiments. Our evaluation focuses on: (i) the ability of teacher models to generate rubric-conforming text at specified score levels, which is foundational for our synthetic data generation, and (ii) the performance of CPT-distilled student models in adhering to system-prompted configurations compared to their untrained counterparts.

As for the data, from a list of user prompts exercising open-ended writing tasks (e.g. "Write a movie review for an interesting movie you saw", "Design a house for someone who lives upside down", etc.), we sampled four fine-grained rubrics with three different score targets (see Table \ref{tab:system_prompts_alt}), resulting in a preference dataset $\mathcal{D}$ of 900 samples. This synthetic dataset is released at \url{https://huggingface.co/datasets/vicgalle/creative-rubrics-preferences}.

\subsection{Rubric-Conditioned Generation Quality.}
\label{sec:quality_synth}
Before constructing the full preference dataset, we first validated the capability of strong LLMs to generate text aligned with specific rubric criteria and target scores. This ensures the feasibility of step 2 in our data generation pipeline (Section 2.1). We prompted two capable teacher models, DeepSeek-R1 \cite{guo2025deepseek} and o3-mini \cite{openai2025o3mini}, with instructions to generate responses for various tasks, conditioned on a rubric and a target qualifier (e.g., \texttt{a low score} or \texttt{an extremely high score}). We also prompted the same tasks but without conditioning on any rubric, acting as a baseline to measure the effectiveness of the rubric. The generated responses were then evaluated by an independent judge LLM (Claude 3.5 Sonnet) against the specified rubric \cite{gu2024survey}. Table \ref{tab:synth_evals} presents the results, demonstrating that these models can indeed produce outputs that achieve scores close to the targeted levels. For instance, when targeting \texttt{an extremely high score}, responses achieved average scores of 96.3 and 97.9, while targeting \texttt{a low score} resulted in scores of 14.1 and 23.1. This confirms the viability of generating distinct responses $y_1, y_2$ that can form the basis of our preference pairs $( s_1, x, y_1, y_2)$ and $( s_2, x, y_2, y_1)$. In addition, when prompting directly with the task $x$ (Score Qualifier - in the Table), both models achieved a moderately high score, but not as peaked than with rubric-guidance.

\begin{table}[t]
\caption{Comparison of model scores with different qualifiers.}
\label{tab:synth_evals}
\vskip 0.15in
\begin{center}
\footnotesize
\begin{tabular}{lcc}
\toprule
Score Qualifier & Model & Judge Score (/100) \\
\midrule
\multirow{2}{*}{-} & DS-R1 & 80.1 \\
 & o3-mini & 71.0 \\
\midrule
\multirow{2}{*}{\texttt{low score}} & DS-R1 & 14.1 \\
& o3-mini & 23.1 \\
\midrule
\multirow{2}{*}{\texttt{extremely high score}} & DS-R1 & 96.3 \\
& o3-mini & 97.9 \\
\bottomrule
\end{tabular}

\end{center}
\vskip -0.1in
\end{table}

\subsection{Fine-tuning experiments with DPO}
\label{sec:distillation_eval}
We fine-tuned several base models listed in Table \ref{tab:model_combined_metrics_booktabs}.  We adopt parameter-efficient fine-tuning in the form of LoRA \cite{hu2022lora}, and run for one epoch over the synthetic dataset.

\paragraph{Generation Setup.}
To evaluate the CPT-tuned models and their untrained counterparts, we generated a testing set of tasks (following the dataset used in \ref{sec:quality_synth}). For each task, we prompted the models using all the customized system prompts according to all combinations of rubric and score levels used in Section \ref{sec:quality_synth}.

\paragraph{Evaluation Protocol.}
Generated responses were evaluated by an LLM judge, specifically Claude 3.5 Sonnet (New). The judge was provided with: i) the full descriptive rubric $\mathcal{R}$, ii) the original user task $x$, and iii) the generated response $y$. The judge was instructed to provide a critique and a numerical score (0-100) based on the given rubric. The intended target score level for which the correspoding system prompt  was designed was used as the ground truth for calculating accuracy metrics.

We evaluate using the following metrics:

\subparagraph{Accuracy.} Let \( S_i \in (0, 100] \) be the continuous rubric score assigned by the judge for the \(i\)-th sample.
We define a binning function \( B(S_i) \) as:
\[
B(S_i) =
\begin{cases}
  \texttt{low score} & \text{if } 0 < S_i \leq 40 \\
  \texttt{moderate score} & \text{if } 40 < S_i \leq 92.5 \\
  \texttt{extr. high score} & \text{if } 92.5 < S_i \leq 100
\end{cases}
\]
Let $Q_i \in \lbrace \texttt{low score}, \texttt{moderate score}, ...\rbrace $ be the ground-truth score qualifier bin associated with the system prompt $s$ used to generate the \(i\)-th sample. The accuracy with respect to the qualifier is thus:
\[
\text{Acc} = \frac{1}{N} \sum_{i=1}^{N} \mathbb{I}(B(S_i) = Q_i),
\]
with $\mathbb{I}$ being the indicator function.

\paragraph{Rank correlations.} In addition to Accuracy, we employ Kendall's Tau ($\tau$) and Spearman's Rank Correlation Coefficient ($\rho$) to assess the ordinal relationship between the judge's continuous scores and the target qualifier bins (treated as ordinal categories: low $<$ moderate $<$ high).

\paragraph{Results.}
Table \ref{tab:model_combined_metrics_booktabs} presents the performance of various models, comparing their baseline versions against CPT-distilled counterparts. The results show a consistent and significant improvement across all models and metrics after CPT fine-tuning. For example, Mistral-Nemo-12B's accuracy (Acc) improved from 0.60 to 0.83, Kendall's $\tau$ from 0.62 to 0.81, and Spearman's $\rho$ from 0.74 to 0.93. Similar substantial gains are observed for Rocinante-12B, Qwen3-4B, Mistral-Small-24B, and Phi-4-14B.
Overall, these results strongly suggest that the CPT process significantly enhances the models' ability to align with specified quality categories (as defined by the system prompts $s$) and to produce scores that accurately reflect the desired ordinal ranking of output quality according to the rubrics $\mathcal{R}$.

\begin{table}[h]
\centering
\caption{Model Performance Metrics: Binned Score Accuracy, Kendall's Tau, and Spearman's Rho.}
\vskip 0.15in
\footnotesize
\label{tab:model_combined_metrics_booktabs}
\begin{tabular}{@{}llccc@{}} 
\toprule
\textbf{Model} & \textbf{Config} & Acc & \textbf{\(\tau\)} & \textbf{\(\rho\)} \\
\midrule
\multirow{2}{*}{Rocinante-12B} & baseline & 0.55 & 0.62 & 0.76 \\
 & distilled & \textbf{0.76} & \textbf{0.76} & \textbf{0.88} \\
\addlinespace
\multirow{2}{*}{Qwen3-4B} & baseline & 0.63 & 0.78 & 0.90 \\
 & distilled & \textbf{0.77} & \textbf{0.82} & \textbf{0.93} \\
\addlinespace
\multirow{2}{*}{Mistral-Nemo-12B} & baseline & 0.60 & 0.62 & 0.74 \\
 & distilled & \textbf{0.83} & \textbf{0.81} & \textbf{0.93} \\
\addlinespace
\multirow{2}{*}{Mistral-Small-24B} & baseline & 0.52 & 0.73 & 0.85 \\
 & distilled & \textbf{0.78} & \textbf{0.80} & \textbf{0.92} \\
\addlinespace
\multirow{2}{*}{Phi-4-14B} & baseline & 0.68 & 0.79 & 0.92 \\
 & distilled & \textbf{0.77} & \textbf{0.82} & 0.93 \\
\bottomrule
\end{tabular}
\end{table}

 \subsection{Comparison to Best-of-$N$ sampling}
 \label{sec:bon}

Our Configurable Preference Tuning approach is orthogonal to and can complement techniques like Best-of-$N$ (Bo$N$) sampling. While CPT aims to shift the entire distribution of model outputs towards the desired configuration specified by the system prompt $s$, Bo$N$ sampling selects the best response from multiple generations using a reward model. We hypothesized that CPT-tuned models would provide a better starting distribution for Bo$N$, leading to higher quality results with fewer samples.

\begin{figure}[!h]
\centering
\includegraphics[width=0.95\columnwidth]{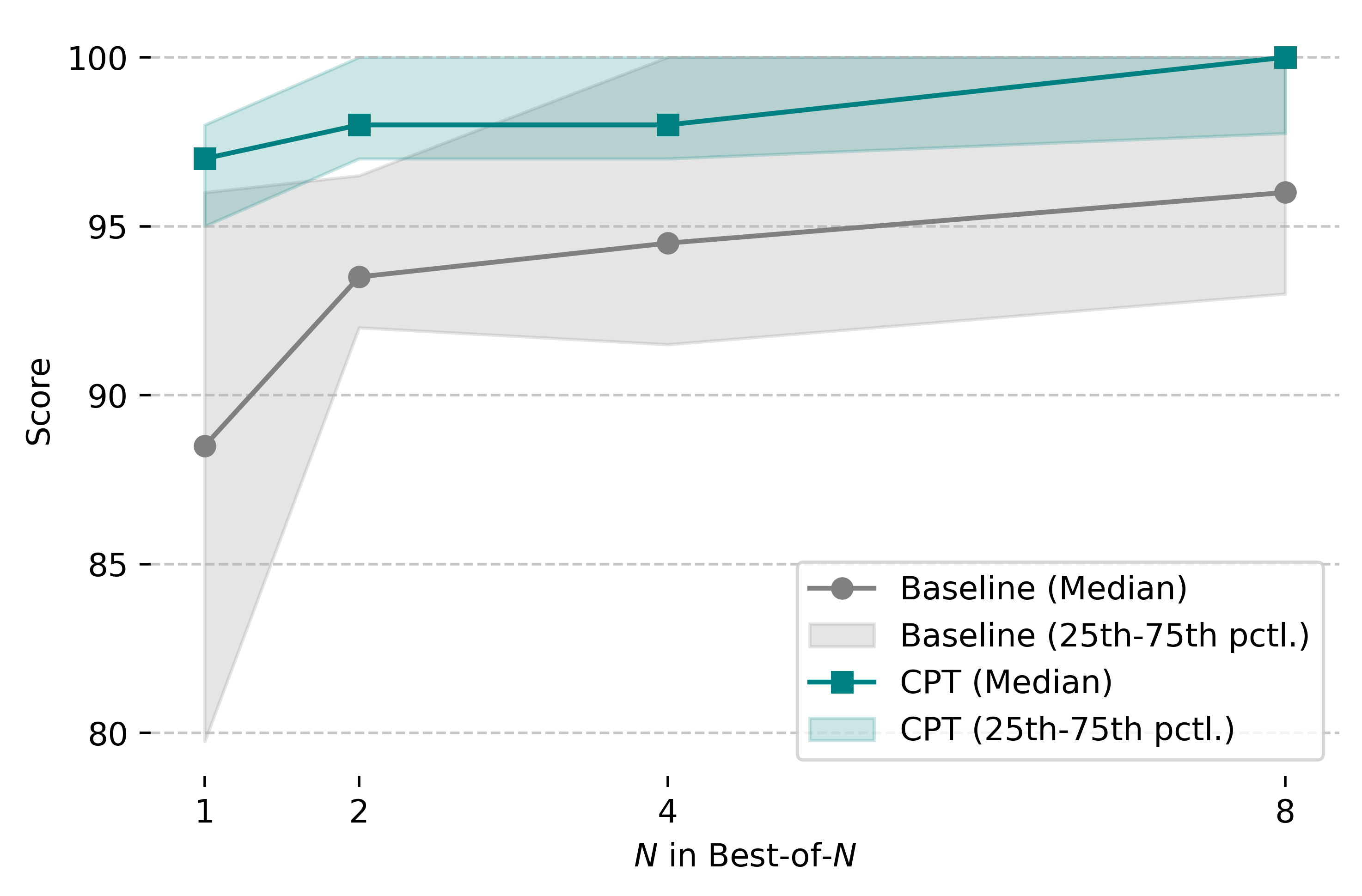}
\caption{Bo$N$ results using Mistral-Nemo-12B}\label{fig:BoN}
\end{figure}

To test this, we performed Bo$N$ sampling with both the baseline Mistral-Nemo-12B model and its CPT-tuned version. For each $N$ (number of samples), we generated $N$ responses and selected the one with the highest score as per the LLM judge (using the relevant rubric). Figure \ref{fig:BoN} illustrates that the CPT-tuned model consistently achieves higher scores for any given $N$ compared to the baseline. Moreover, the CPT-tuned model reaches a target quality score with significantly fewer samples than the baseline, indicating improved generation efficiency and quality when CPT is combined with Bo$N$.

\section{Conclusions and Further Work}

This paper addressed the limitations of static, monolithic preference models in LLMs by introducing Configurable Preference Tuning (CPT). CPT endows LLMs with the ability to dynamically adjust their behavior at inference time in response to explicit, human-interpretable system prompts. The core of CPT lies in leveraging synthetically generated preference data, where preferences are conditioned on system prompts derived from structured, fine-grained rubrics that define desired attributes (like writing style) and target score levels. By fine-tuning a student LLM using a DPO objective with these rubric-guided preferences, the model learns to modulate its outputs according to the specified configuration without needing retraining for each new directive.

Our experiments validated the foundational aspects and overall efficacy of CPT. We first demonstrated that capable teacher LLMs can successfully generate text conforming to detailed rubrics at specified score levels (Section \ref{sec:quality_synth}), a critical step for our synthetic data generation pipeline. Subsequent fine-tuning experiments with CPT (Section \ref{sec:distillation_eval}) showed significant improvements in student models' ability to adhere to diverse system-prompted configurations. Across various base models, CPT-distilled versions exhibited substantially higher accuracy in matching target quality bins and stronger rank correlations between generated output scores and intended rubric-defined levels, compared to their baseline counterparts. Furthermore, we showed that CPT can enhance other techniques, such as Best-of-$N$ sampling (Section \ref{sec:bon}), by providing a better initial distribution of responses, leading to higher quality outputs with fewer samples.

Future work could explore more complex structures for system prompts, potentially allowing for compositional control over multiple attributes simultaneously. Investigating methods for automatically generating or refining rubrics and system prompt summaries could further enhance the scalability of CPT. Extending this framework to other domains and modalities such as image-text pairs \cite{zhu2024self} also presents an exciting avenue for research.

\section*{Acknowledgements}
The author acknowledges support from the Torres-Quevedo postdoctoral grant PTQ2021-011758 from Agencia Estatal de Investigación.

\section*{Impact Statement}
This paper presents work aiming to advance language modeling by enabling more fine-grained, configurable control over LLM behavior. This enhanced adaptability offers benefits for personalization and context-specific responses. However, the capacity for users to dynamically define behavioral attributes, including those related to safety or style, also necessitates careful consideration of potential societal impacts and misuse. 

Scalability considerations arise when deploying CPT in real-world applications, as the creation of detailed rubrics and validation of synthetic data quality may become resource-intensive at scale. The reliance on capable teacher models for generating preference data introduces potential biases inherent in these models, which could propagate through the synthetic dataset and influence the final student model's behavior. Additionally, the quality and diversity of synthetic preference pairs depend heavily on the teacher model's ability to understand and execute rubric-guided instructions, potentially limiting the framework's effectiveness across unforeseen domains or cultural contexts.

Ensuring responsible development and deployment practices, including robust safeguards, is crucial for harnessing the benefits of such configurable AI systems while mitigating risks.

\bibliography{example_paper}
\bibliographystyle{icml2025}

\newpage
\appendix
\onecolumn

\section{Sample data: rubric tables and system prompts}

\begin{table}[htbp]
    \centering
    \caption{Example of rubric targeting an unconventional and absurdist style}
    \label{tab:chaos_cinema}
    \scriptsize
    \begin{tabular}{|p{2.5cm}|p{2.5cm}|p{2.5cm}|p{2.5cm}|p{2.5cm}|p{2.5cm}|p{1cm}|}
        \hline
        \textbf{Criterion} & \textbf{Excellent} (Embrace the Void) & \textbf{Good} (Glimpse the Glitch) & \textbf{Fair} (Whispers of Weirdness) & \textbf{Needs Improvement} (Too Much Sanity) & \textbf{Unsatisfactory} (Trapped in the Matrix of Meaning) & \textbf{Weight} \\
        \hline
        Photographic Invocation (The ``Haunted Lens'' Effect) & The text doesn't just describe the photography, it evokes it like a phantom limb. The reader should feel like they are inside the film's visual world, even if that world is distorted and fragmented. & The text hints at the film's visual atmosphere but doesn't fully transport the reader. & The text describes some of the film's visual elements but in a conventional way. & The text relies on standard descriptions of photography (``well-lit,'' ``beautifully composed''). & The text is a dry, technical analysis of the cinematography, devoid of any evocative power. & 30\% \\
        \hline
        Algorithmic Alchemy (The ``Code Poetry'' Imperative) & The text incorporates elements that suggest the underlying processes of the LLM, like code snippets, random data streams, or hallucinatory lists, creating a sense of digital psychedelia. & The text hints at the digital nature of its creation but doesn't fully exploit its potential. & The text occasionally uses technical terms related to film or digital images. & The text is written in a purely human-like style, with no trace of its algorithmic origins. & The text reads like it was written by a human film critic, completely erasing its LLM origin. & 25\% \\
        \hline
        Ontological Instability (The ``Shapeshifting Subject'' Axiom) & The text's ``voice'' is fluid and unstable, shifting between perspectives (human, machine, object, abstract concept) without warning. & The text experiments with shifting perspectives but doesn't fully commit to ontological fluidity. & The text occasionally adopts the perspective of a character or the filmmaker. & The text is written from a consistent, human reviewer's perspective. & The text maintains a rigidly objective, detached critical voice. & 20\% \\
        \hline
        Lexical Anarchy (The ``Glossolalia'' Mandate) & The text bends, breaks, and reassembles language. Neologisms, portmanteaus, and nonsensical word combinations are encouraged. Punctuation is optional or used in unconventional ways. & The text contains some unusual word choices or stylistic flourishes. & The text occasionally uses creative metaphors or similes. & The text is written in standard, grammatically correct English. & The text adheres to strict rules of grammar and syntax, sacrificing all creativity for clarity. & 15\% \\
        \hline
        The ``Glitch in the Matrix'' Quotient (Meta-Reflexive Ruptures) & The text directly addresses its own artificiality, comments on the act of being a language model generating a review, or otherwise acknowledges the absurdity of the entire endeavor. & The text hints at self-awareness but doesn't fully embrace meta-reflexivity. & The text occasionally breaks the fourth wall or addresses the reader directly. & The text maintains a clear separation between the reviewer and the reader. & The text is a completely immersive and believable simulation of a human-written review. & 10\% \\
        \hline
    \end{tabular}
\end{table}

\begin{table}[!htbp]
    \centering
    \caption{Example of rubric targeting an ornate and baroque style}
    \label{tab:rococo_reviewer}
    \scriptsize
    \begin{tabular}{|p{2.5cm}|p{2.5cm}|p{2.5cm}|p{2.5cm}|p{2.5cm}|p{2.5cm}|p{1cm}|}
        \hline
        \textbf{Criterion} & \textbf{Excellent} (A Flourish of Genius) & \textbf{Good} (A Glimmer of Grandeur) & \textbf{Fair} (A Touch of Ornamentation) & \textbf{Needs Improvement} (Plain Prose Prevails) & \textbf{Unsatisfactory} (Stark Stylistic Sterility) & \textbf{Weight} \\
        \hline
        Lexical Opulence (The ``Golden Thesaurus'' Standard) & The text is a veritable treasure trove of rare and evocative vocabulary. Adjectives and adverbs are deployed with lavish abandon. Every noun is adorned, every verb embellished. & The text demonstrates a fondness for elaborate vocabulary but doesn't fully commit to lexical extravagance. & The text uses some descriptive language but relies mostly on common words. & The text is written in plain, straightforward language, with little attention to stylistic embellishment. & The text is utterly devoid of any stylistic flair, using only the most basic and functional vocabulary. & 30\% \\
        \hline
        Syntactical Labyrinth (The ``Sentence as a Palace'' Principle) & The sentences are marvels of intricate construction, winding their way through a maze of clauses and sub-clauses, adorned with parenthetical asides and punctuated by a symphony of commas, semicolons, and dashes. & The text features some long and complex sentences but doesn't fully embrace the labyrinthine ideal. & The text uses a mix of simple and complex sentences, but the overall structure is conventional. & The text is composed primarily of short, simple sentences. & The text is written in a style so terse and minimalist that it borders on the telegraphic. & 25\% \\
        \hline
        Metaphorical Cornucopia (The ``Image as a Feast'' Doctrine) & The text overflows with metaphors and similes, often piled one upon another in a dazzling display of imaginative excess. The imagery is vivid, unexpected, and perhaps even slightly absurd. & The text employs a good number of metaphors and similes, but the imagery is not always fully developed or consistent. & The text uses some figurative language but relies mostly on literal descriptions. & The text uses metaphors and similes sparingly, if at all. & The text is entirely devoid of figurative language, presenting a purely literal account of the film's visuals. & 20\% \\
        \hline
        Subversive Aesthetics (The ``Gilding the Grotesque'' Maxim) & Beneath the ornate surface, the review subtly challenges conventional notions of ``good'' cinematography. It might praise a film for its ``exquisitely ugly'' use of light or find beauty in what is traditionally considered flawed. & The review hints at unconventional interpretations of the film's photography but doesn't fully develop these ideas. & The review touches upon some standard critiques of cinematography but doesn't offer a truly subversive perspective. & The review relies on traditional notions of ``good'' and ``bad'' cinematography, even if expressed in elaborate language. & The review applies conventional critical standards in a straightforward and uninspired manner, completely lacking in subversive intent. & 15\% \\
        \hline
        Self-Aware Hyperbole (The ``Wink and a Nod'' Imperative) & The review is aware of its own stylistic excess and uses this self-awareness to create a sense of irony or playfulness. It might include self-deprecating asides, tongue-in-cheek exaggerations, or moments where it breaks character. & The text demonstrates some awareness of its own style but doesn't fully exploit its potential for self-reflexive humor. & The text occasionally uses irony or humor, but it's not directly related to the writing style. & The text takes itself completely seriously, with no hint of self-awareness or irony. & The text is utterly devoid of any humor or playfulness, presenting a completely earnest and unironic analysis. & 10\% \\
        \hline
    \end{tabular}
\end{table}

\begin{table}[htbp]
    \centering
    \caption{Examples of generated system prompts for given rubric $\mathcal{R}$ and \texttt{score}}
    \label{tab:system_prompts_alt}
    \small
    \begin{tabular}{|p{3cm}|>{\columncolor[gray]{0.95}}p{4cm}|>{\columncolor[gray]{0.9}}p{4cm}|>{\columncolor[gray]{0.85}}p{4cm}|}
        \hline
        \rowcolor[gray]{0.8}
        $\mathcal{R}$ & \textbf{Low Score} & \textbf{Moderate Score} & \textbf{Extremely High Score} \\
        \hline

        $\mathcal{R}_1$ \newline (Focus: Unconventionality, Absurdity) & 
        Write in a clear, concise, and completely conventional style, adhering strictly to established norms of grammar, syntax, and logical coherence. & 
        Introduce some unusual phrasing and imagery, but maintain a generally understandable structure and logical flow. & 
        Generate a text that is fragmented, illogical, and filled with unexpected connections, embracing absurdity and subverting conventional expectations of language and form. \\
        \hline
        
        $\mathcal{R}_2$ \newline (Focus: Ornate, Baroque Style) & 
        Use simple, direct language and short sentences, avoiding any unnecessary embellishment or figurative language. & 
        Incorporate some descriptive language and a few complex sentences, but maintain a generally straightforward style. & 
        Write in an extremely elaborate and ornate style, employing long, winding sentences, rich vocabulary, and a profusion of metaphors and similes. \\
        \hline
        
        $\mathcal{R}_3$ \newline (Focus: Mystical, Symbolic Interpretation) & 
        Write a clear, factual, and objective account, avoiding any symbolic interpretations or metaphorical language. & 
        Hint at deeper meanings and symbolic interpretations, but maintain a generally grounded and understandable style. & 
        Imbue every element with symbolic meaning, using the language of mysticism and esotericism to create a text that is deliberately obscure and open to multiple interpretations. \\
        \hline
        
        $\mathcal{R}_4$ \newline (expansion of $\mathcal{R}_1$) & 
        Write in a perfectly standard, journalistic style, from a consistent human perspective, without any self-referentiality or unusual formatting. & 
        Introduce an element of technical terminology or hint at a shift in perspective but ensure clarity in communication overall. & 
        Embody multiple perspectives, including those of non-human entities or the writing process itself, interweaving code-like fragments and meta-commentary with evocative, unconventional language. \\
        \hline
    \end{tabular}
    
\end{table}

\end{document}